\documentclass{article}

\usepackage{booktabs}

\usepackage[hidelinks]{hyperref}

\usepackage{adjustbox}
\usepackage{graphicx}
\usepackage{amsmath, amssymb, amsthm} 
\usepackage{gensymb}
\usepackage{subfig}
\usepackage{tikz}
\usepackage{pgfplots}

\usepackage{diagbox}
\usepackage{multirow}
\usepackage{caption}

\usepackage{fancyhdr}
\usepackage{array}
\usepackage{wrapfig}
\usepackage{float}
\usepackage{url}

\newcolumntype{M}[1]{>{\centering\arraybackslash}m{#1}}

\begin{document}

\title{Seasonal Fire Prediction using Spatio-Temporal Deep Neural Networks}

\date{}

\author{
    Dimitrios Michail\thanks{Harokopio University of Athens, Greece. \{michail, epanagiotou, cdavalas\}@hua.gr} \and 
    Lefki-Ioanna Panagiotou\footnotemark[1] \and
    Charalampos Davalas\footnotemark[1] \and 
    Ioannis Prapas \thanks{OrionLab, National Technical University \& National Observatory of Athens, Greece. \{iprapas, skondylatos, bountos, ipapoutsis\}@noa.gr} \and
    Spyros Kondylatos\footnotemark[2] \and 
    Nikolaos Ioannis Bountos\footnotemark[1] \footnotemark[2] \and 
    Ioannis Papoutsis \footnotemark[2]
}

\maketitle

\begin{abstract}
With climate change expected to exacerbate fire weather conditions, the accurate anticipation of wildfires on a global scale becomes increasingly crucial for disaster mitigation. In this study, we utilize SeasFire, a comprehensive global wildfire dataset with climate, vegetation, oceanic indices, and human-related variables, to enable seasonal wildfire forecasting with machine learning. 
For the predictive analysis, we train deep learning models with different architectures that capture the spatio-temporal context leading to wildfires. 
Our investigation focuses on assessing the effectiveness of these models in predicting the presence of burned areas at varying forecasting time horizons globally, extending up to six months into the future, and on how different spatial or/and temporal context affects the performance of the models. Our findings demonstrate the great potential of deep learning models in seasonal fire forecasting; longer input time-series leads to more robust
predictions across varying forecasting horizons, while 
integrating spatial information to capture wildfire spatio-temporal dynamics boosts performance. Finally, our results hint that in order to enhance performance at longer forecasting horizons, a larger receptive field spatially needs to be considered. 
\end{abstract}

\definecolor{myblue}{RGB}{0, 102, 204} 
\definecolor{mygreen}{RGB}{34, 139, 34}
\definecolor{myorange}{RGB}{255, 128, 0}
\definecolor{mycyan}{RGB}{0, 255, 255}  
\definecolor{mymagenta}{RGB}{255, 0, 255}  
\definecolor{mypurple}{RGB}{128, 0, 128} 
\definecolor{myyellow}{RGB}{255, 255, 0} 
\definecolor{mygray}{RGB}{128, 128, 128} 









\section{Introduction}\label{sec:intro}

Fire plays a pivotal role in the Earth system, significantly influencing ecosystems worldwide. While 
traditionally viewed as a carbon-neutral process over the long term, climate change disrupts this balance
through the intensification of fire weather conditions, leading to a rise in global fire activity~\cite{jones2022global}. 
A feedback loop is created when fires encroach into evergreen forest regions, posing a threat to their
role as carbon sinks. This situation triggers the release of stored carbon into the atmosphere, further exacerbating 
climate change~\cite{flannigan2009implications}.
Moreover, wildfires represent a critical natural hazard with far-reaching consequences for
societies worldwide, resulting in loss of life, property, infrastructure, and ecosystem
services~\cite{pettinari2020fire}.
Hence, it is imperative to improve our understanding and forecasting capabilities of wildfire phenomena within the Earth system. By doing so, we can formulate effective strategies to mitigate the adverse impacts of wildfires and climate change. 

In this study, our objective is to improve our capacity for wildfire prediction, ultimately contributing to the broader mission of safeguarding ecosystems and societies in the face of evolving climate conditions. We accomplish this by applying and testing different deep learning architectures, e.g. Gated Recurrent Units (GRU), Convolutional Long Short-Term Memory (Conv-LSTM), as well as temporally enabled Graph Neural Networks (TGNN) to environmental data.
For all these architectures, we study the impact of the spatio-tempral context on the predictive skill of our models, 
through a series of experiments with different time-series lengths, spatial neighborhoods, and prediction horizons.
All source code can be found at \url{https://github.com/seasFire/seasfire-ml}.

\section{Related Work}

Wildfires are notoriously hard to model due to the non-linear interactions between the different Earth system processes that affect them \cite{hantson2016status}. Weather, vegetation and humans interact in evolving ways that contribute in the expansion or suppression of wildfires. Reichstein et al. \cite{reichstein2019deep} propose Deep Learning as a method to learn in a  data-driven way these complex spatiotemporal interactions that influence wildfires. Several recent studies have used deep learning for wildfire-related use cases \cite{jain_review_2020}. For short-term daily predictions the temporal context is mostly enough, with spatiotemporal models offering little to no advantage \cite{prapas2021deep, kondylatos2022wildfire}. For longer-term predictions, i.e. on subseasonal to seasonal scales, very few works have studied the effect of spatial and temporal context. 
Joshi et al. \cite{joshi2021improving} use monthly aggregated input to predict burned area using multi-layer neural networks. Li et al. \cite{li2023attentionfire_v1} includes the temporal aspect of the input using a temporal attention network on time-series of fire driver variables to predict the burned area over the tropics. Prapas et al. \cite{prapas2022deep} use temporal snapshots of the fire drivers to predict future burned area patterns, defining burned area forecasting as a segmentation task and demonstrating skillful forecasts even two months in advance. An expansion of this \cite{Prapas_2023_ICCV}, leverages teleconnection indices and global input in conjunction with those snapshots, which helps improve long-term skill, but ignores the temporal component of the input variables. 
Finally, Zhao et al. \cite{zhao2024causal} integrate causality with GNNs to explicitly model the causal mechanism among complex variables via graph learning, and test their models in the European boreal and Mediterranean biome.

In this work, we examine the performance of various Deep Learning models leveraging the spatiotemporal context of varying size to predict the burned area pattern weeks to months in advance, in global scale and at $1^{\circ}$ spatial resolution. We do an extensive study to identify optimal sizes for the spatial and temporal context and benchmark different types of architectures including GRUs, temporal GNNs and ConvLSTMs. 

\section{Seasonal Fire Forecasting}

\begin{figure*}[t]
    \centering
    \includegraphics[width=0.95\columnwidth, keepaspectratio]{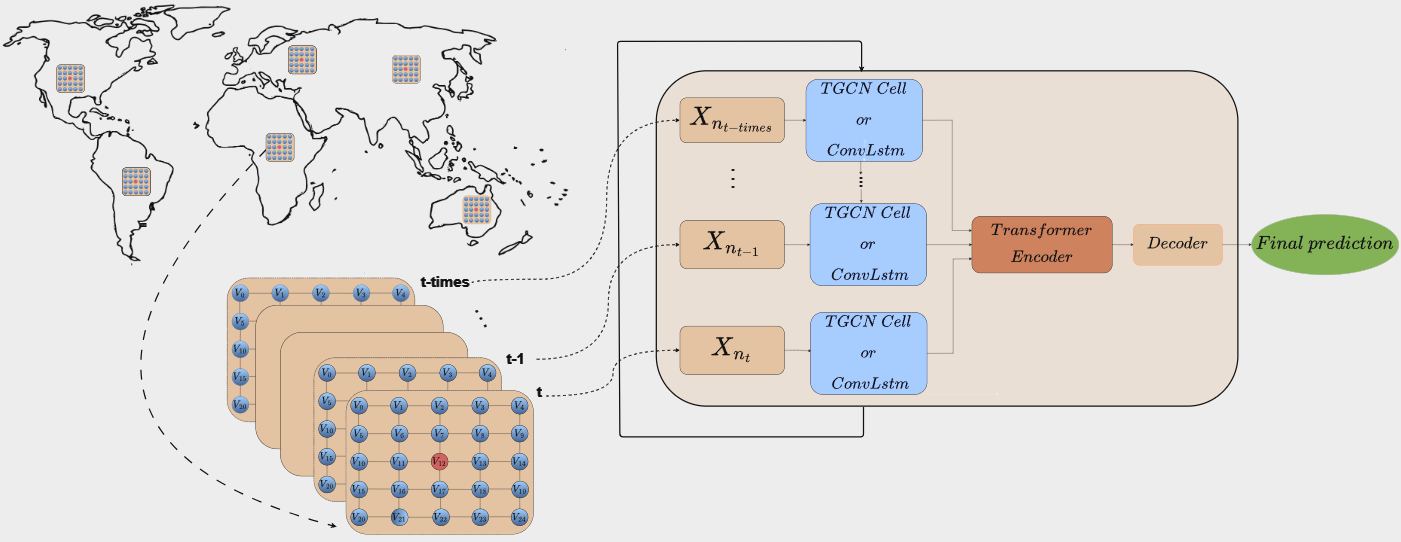} 
    \caption{High-level representation of our architecture. The $5\times5\times$ $t$ grid (lower left) represents the structure of the model input. The central (red) vertex is the location in which we try to predict the fire hazard probability. Time $t$ is the time right before the time of the prediction and time $t-times$ is the start time of the time series. $X_n$ stands for the $n\textsuperscript{th}$ sub-graph in the dataset.}
    \label{fig:high_level_representation}
\end{figure*}

\subsection{Dataset}\label{sec:dataset}

\begin{table}[t]
\centering
\begin{tabular}{cc}
\toprule
\textbf{Fullname} & \textbf{Variable name}\\
\midrule
Mean sea level pressure & mslp \\
Total precipitation & tp \\
Vapor Pressure Deficit & vpd \\
Sea surface temperature & sst \\
Temperature at 2 meters - Mean & t2m\_mean \\
Surface solar radiation downwards & ssrd \\
Volumetric soil water level 1 & swvl1 \\
Land surface temperature at day & lst\_day \\
Normalized Difference Vegetation Index & ndvi \\
Population density & pop\_dens \\
Latitude (sin/cos) &  cube coordinates \\
Longitude (sin/cos) & cube coordinates  \\
\midrule
\textbf{Target} & \textbf{Variable name}\\    
\midrule
Burned Areas (as binary) & gwis\_ba\\
\bottomrule
\end{tabular}

\caption{Basic variables (10 input and 2 positional) used for predictions.
Target variable represents burned area in hectares (converted to binary).}
\label{tab:variables}
\end{table}

The SeasFire Datacube~\cite{karasante2023seasfire} is an analysis-ready, open-access datacube fit for wildfire forecasting at different spatiotemporal scales. 
It spans over 21 years (2001-2021) with an 8-day temporal and a $1^{\circ}$ spatial resolution (there is also a $0.25^{\circ}$ version available). 
The dataset provides a comprehensive coverage of atmospheric, climatological, vegetation and socioeconomic factors influencing wildfires and contains
58 variables in total, along with target variables,
 such as burned areas, fire radiative power, and carbon emissions from wildfires. 

In this work, we utilized 10 different variables (see Table~\ref{tab:variables}) in order to predict 
the existence of fire in different time horizons. In addition and when needed, we also used simple positional encodings by augmenting
the feature vector with the actual cube coordinates, i.e. sin/cos of latitude and longitude. 
All variables, except the cube coordinates, were standardized before use. 

\subsection{Problem formulation and methodology}

We view the problem as binary classification at a particular location of the cube
and a particular timestamp. Thus, given a triplet $(\phi_c, \lambda_c, t_c)$ of latitude, longitude and 8-day period we
predict whether a fire will occur or not at that particular location in time and space. As input we use a timeseries for
each variable of length $ts \in \{6, 12, 24, 36\}$ timesteps, where each timestep corresponds to a single 8-day period.
As target variable  
we predict the presence of burned areas at a future timestep $t+h$ with different 
values of $h \in \{1, 2, 4, 8, 12, 16, 20, 24\}$.

\subsubsection{Naive Forecasting}\label{sec:naive}

When dealing with seasonal time 
series, the naive seasonal forecasting method uses the observed values in the 
same period of the previous seasons~\cite{hyndman2018forecasting}. 
Our dataset encompasses multiple years, enabling the utilization of various baseline methods based on past data. We employ two approaches: Firstly, for forecasting fire occurrence in a specific 8-day period at a given location, we examine historical data to determine if fires were previously recorded in that period at the location. If so, a fire is predicted; otherwise, no fire is predicted. Secondly, using a majority rule, we predict fire occurrence at a location for a specific 8-day period if the number of previous years with fires in that period surpasses the number of years without fires. This majority baseline approach aligns effectively with the binary classification task.

\begin{figure}[th]
    \centering
    \includegraphics[width=0.8\columnwidth]{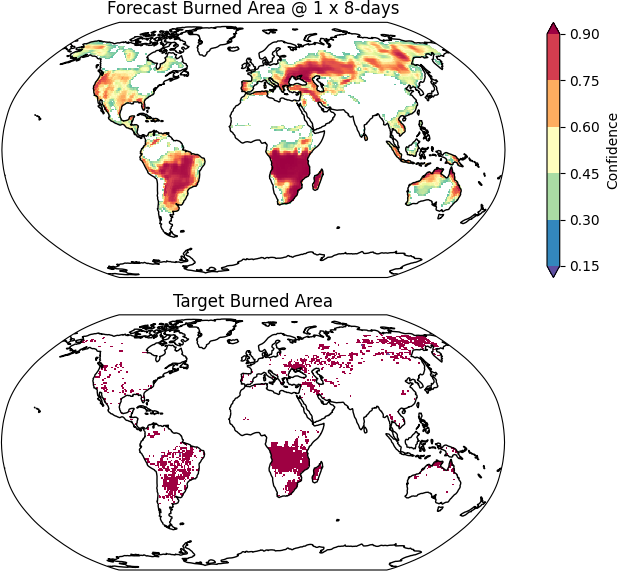} 
    \caption{Target variable and prediction of the best global model, using a time-series of length 36, at 07-19-2020.
    Target is one period (8-days) ahead.  Confidence score is the sigmoid output.}
    \label{fig:example:prediction}
\end{figure}

\subsubsection{Recurrent Neural Networks}

Recurrent Neural Networks~\cite{medsker2001recurrent} are a natural choice when trying to capture temporal dependencies.
We utilize a very simple architecture comprising of Gated Recurrent Units~\cite{cho2014learning}. The neural network is
comprised of two GRU layers with 64 hidden channels in each layer. A dropout layer with a probability of 0.1 exists between
the two GRU layers. As the task is binary classification, a final linear layer reduces the representation into a single
prediction and a sigmoid function outputs the final prediction.

\subsubsection{Convolutional Recurrent Neural Networks}\label{sec:conv-lstm}

To capture both temporal and spatial dependencies at the same time, we utilize a convolutional
LSTM network~\cite{shi2015convolutional}. In such a network all inputs, cell outputs, hidden states 
and gates are 3D tensors where the last two dimensions are spatial dimensions (rows and columns).
The future state of a certain cell is computed from the input and the past states of its local 
neighbors. This is achieved by using convolution operators in different transitions such as the
state-to-state and/or input-to-state. We utilize a slightly different network from the one
in~\cite{shi2015convolutional} 
which is implemented using the following equations:
\begin{align*}
   i_t & = \sigma(W_{xi} * X_t + W_{hi} * h_{t-1} + b_i)\\
   f_t & = \sigma(W_{xf} * X_t + W_{hf} * h_{t-1} + b_f)\\
   o_t & = \sigma(W_{xo} * X_t + W_{ho} * h_{t-1} + b_o)\\
   g_t & = tanh(W_{xg} * X_t + W_{hg} * h_{t-1} + b_g)\\
   c_t & = f_t \odot c_{t-1} + i_t \odot g_t\\
   h_t & = o_t \odot tanh(c_t)
\end{align*}
Here $*$ denotes the convolution operator and $\odot$ point-wise 

Given $(\phi_c, \lambda_c)$ for a particular location of interest and a particular time $t$ we generate
a 3D tensor $X_t$ whose 
last two dimensions are the number of rows and columns of a $(2r+1) \times (2r+1)$ spatial grid. The 
first dimension is the number of features.
The grid is centered at our location of interest
while the remaining locations represent $(\phi_c \pm i \cdot 1^{\circ}, \lambda_c \pm i \cdot 1^{\circ})$
for $i \in \{1, \ldots, r-1\}$. We feed $X_t$ to the model for each timestep $t$ in order to compute 
the hidden representation $h_t$.
The final result is obtained by utilizing MultiLayer Perceptron (MLP) layers in order to gradually aggregate the information into a single
prediction.

\subsubsection{Temporal Graph Convolutional Network}\label{sec:local}

A slightly different approach is to utilize 
a temporal graph convolutional network model (T-GCN) which combines a graph convolutional network (GNN) and gated
recurrent units~\cite{zhao2019t}. Figure \ref{fig:high_level_representation} shows a high-level representation of the proposed architecture. The GNN is used to capture the topological structure at the location that we are
trying to forecast and the GRU captures the temporal dynamics of the environmental information. 

Given $(\phi_c, \lambda_c)$ for a particular location of interest we generate a small grid
graph centered around it depending on a radius $r > 0$ hyper-parameter. Given $r$ the graph contains 
$(2r+1) \times (2r+1)$ vertices arranged in a spatial grid. The center vertex is our location of interest
while the remaining vertices represent locations at $(\phi_c \pm i \cdot 1^{\circ}, \lambda_c \pm i \cdot 1^{\circ})$
for $i \in \{1, \ldots, r-1\}$. For each of these vertices we connect it to its $k$ nearest neighbors where 
again $k$ is a hyper-parameter.

Given the adjacency matrix $A$ of our grid based graph and the feature matrix $X$, the GCN model
can be written as: 
\[
    H^{(l+1)} = \sigma(\tilde{D}^{-\frac{1}{2}}\hat{A}\tilde{D}^{-\frac{1}{2}}H^{(l)}W^{(l)}),
\]
where $\hat{A} = A + I$ is the adjacency matrix of the graph after adding self-loops, 
$\tilde{D}$ is the degree matrix of the graph with the self-loops, 
$H^{(l)}$ is the output of the $l$-th layer, $W^{(l)}$ is the matrix with the trainable parameters
of the $l$-th layer, and $\sigma(\cdot)$ represents the non-linear activation function. 
A $2$-layer GCN model~\cite{kipf2016semi} can be expressed as 
\[ 
    f(A, X) = \sigma(\hat{A} \cdot Relu(\hat{A} \cdot X \cdot W^{(0)}) \cdot W^{(1)}),
\]
where $\hat{A}$ is the normalized adjacency matrix and can be computed in a preprocessing step. 
Matrices $W^{(0)} \in \mathbb{R}^{d \times h_1}$ and $W^{(1)} \in \mathbb{R}^{h_1 \times h_2}$ are the 
trainable matrices for the two layers where $d$ is the number of input features and $h_1, h_2$ the 
length of the hidden representations. In our experiments we used $64$ hidden channels per GCN layer, 
i.e. $h_1=h_2=64$.

The output of the second layer of the GCN is a hidden representation $h \in \mathbb{R}^{h_2}$ for each
vertex of the graph. The T-GCN cell is a GRU cell which given a vertex hidden representation $h_{t-1}$
at timestep $t-1$, produces the output $h_t$ at time $t$, given the following equations
\begin{align*}
    u_t & = \sigma(W_u \left( f_u(A, X_t) \oplus h_{t-1} \right) + b_u)\\
    r_t & = \sigma(W_r \left( f_r(A, X_t) \oplus h_{t-1} \right) + b_r)\\
    c_t & = tanh( W_c \left( f_c(A, X_t) \oplus (r_t \odot h_{t-1}) \right) + b_c  )\\
    h_t & = u_t \odot h_{t-1} + (1-u_t) \odot c_t
\end{align*}
In the above, $\oplus$ denotes vector concatenation and $\odot$ point-wise multiplication. Note that 
the model uses different parameters for each of the three GCN models.

The final result is obtained by aggregating the final hidden representations of the T-GCN cells 
at each graph vertex.
Recent works~\cite{corso2020principal, li2020deepergcn} show that using multiple aggregations and
learnable aggregations can potentially provide substantial improvements.
Therefore, we first apply multi-head self-attention~\cite{vaswani_attention_2017} 
and finally use MLP layers in order to gradually aggregate the information of all vertices 
to a representation for the entire graph and finally output a single prediction. In our experiments we utilized 
an attention mechanism with 4 heads and 256 hidden dimension. 

\section{Experiments}\label{sec:experiments}

Our methodology can be directly applied to any region of the world, producing different models per region. 
For our forecasting task, our split is time-based, using years 2002-2017, 2018 and 2019 for train, validation and test respectively.
Figure~\ref{fig:example:prediction} shows the ground truth and the model's predictions for a specific timestamp, trained to forecast 8 days ahead globally.

    Due to our highly imbalanced dataset, the model’s performance is evaluated using the Area
    Under the Precision-Recall Curve (AUPRC)~\cite{davis2006relationship}. 
    We calculate the metric using the average precision score which 
    summarizes a precision-recall curve as a weighted mean of precisions at each threshold, 
    with the difference in recall from the previous threshold as weight
    \[
        AP = \underset{t}{\sum} (R_t - R_{t-1}) \cdot P_t, 
    \]
    where $P_t, R_t$ is the respective precision and recall at threshold index $t$. 

    Models are trained for 100 epochs. We use binary cross entropy loss and the SGD optimizer. 
    The initial learning rate is set to $0.01$ and it adjusted using SGDR~\cite{DBLP:conf/iclr/LoshchilovH17}. 
    The schedule contains two cycles with 25 and 75 epochs respectively. A weight decay factor of 0.001 is 
    used while training all models.

    \subsection{Global model}

The naive forecasting baseline models discussed in Section~\ref{sec:naive} 
    exhibit $0.35$ and $0.39$ AUPRC. The first baseline checks whether any previous year contains 
    some fire in the same period while the second whether a majority of the previous years
    contains fires. 

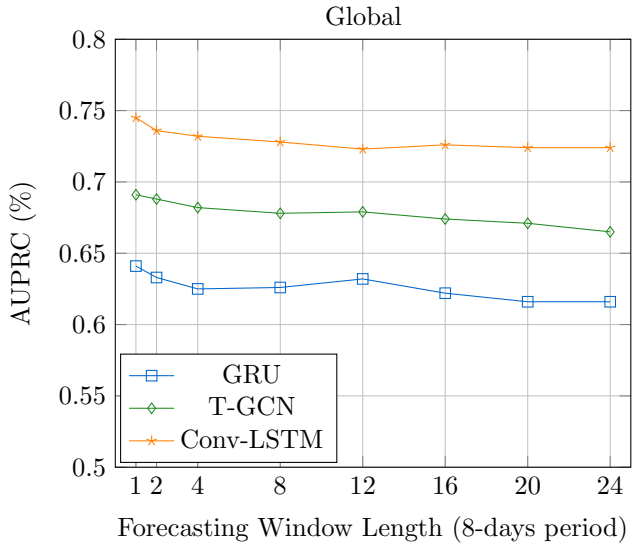
\begin{figure}[th]
        \centering
        \begin{tikzpicture}
            \begin{axis}[
                xlabel={Forecasting Window Length (8-days period)},
                ylabel={AUPRC (\%)},
                legend style={at={(0.22,0.01)},anchor=south},
                xtick={1,2,4,8,12,16,20,24},
                ytick={0.4, 0.45,0.5, 0.55,0.6, 0.65,0.7, 0.75, 0.8, 0.85,0.9,1.0},
                xmin=0,
                xmax=25,
                ymin=0.5, 
                ymax=0.8,
                grid=both,
                major grid style={line width=.1pt,draw=mygray!50},
                minor tick num=1,
                every major grid/.style={line width=.1pt,draw=mygray!50},
            ]
            \addplot[mark=square, color=myblue] table [x=x, y=ts36] {gru-globe.dat};
            \addplot[style={solid}, mark=diamond, color=mygreen] table [x=x, y=ts36] {tgcn-local-radius5-globe.dat};
            \addplot[mark=star, color=myorange] table [x=x, y=ts36] {conv-lstm-local-radius5-globe.dat};

            \legend{GRU, T-GCN, Conv-LSTM}

            \end{axis}
            \node at (3.3,6.0) {Global};
        \end{tikzpicture}
        \caption{Performance of GRU, Conv-LSTM and T-GCN models at a global scale with timeseries length
        $36$ for 
        different forecasting windows.}\label{fig:all:ts36:globe:auprc}
    \end{figure}

    Figure~\ref{fig:all:ts36:globe:auprc} presents the best model results when predicting at a global scale. 
    Different GRU, Conv-LSTM and T-GCN models were trained for each different prediction time horizon, 
    resulting in 24 distinct models. Both the Conv-LSTM and the T-GCN model use a radius $r=5$ resulting 
    in a spatial grid of size $11 \times 11$ around the area of interest. The timeseries for all models has 
    a length of 36 8-day periods.

    Conv-LSTM starts from 0.74 for the next period prediction, showly dropping to 0.72 in case of 24 periods. 
    The stability of the prediction, even for large prediction horizons, is mostly due to the 
    large length of the timeseries that is fed to the model.  
    T-GCN starts from 0.69 for the next period prediction which slowly drops to 0.66 when predicting after 
    24 periods. Similarly, GRU starts from 0.64 down to 0.61.

    \subsection{Timeseries length and spatio-temporal context}

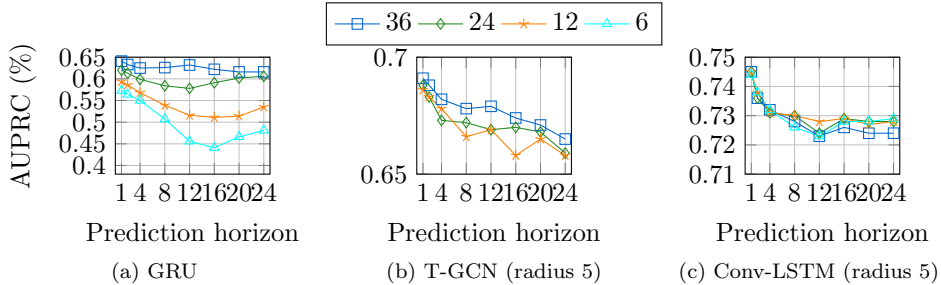
\begin{figure*}[t]
        \centering
        \subfloat[GRU\label{fig:global:timeseries:gru}]{
            \begin{tikzpicture}
                \begin{axis}[
                    xlabel={Prediction horizon},
                    ylabel={AUPRC (\%)},
                    xtick={1,4,8,12,16,20,24},
                    ytick={0.4, 0.45,0.5, 0.55,0.6, 0.65,0.7,0.8,0.9,1.0},
                    xmin=0,
                    xmax=25,
                    ymin=0.38, 
                    ymax=0.65,
                    grid=both,
                    major grid style={line width=.1pt,draw=mygray!50},
                    minor tick num=1,
                    every major grid/.style={line width=.1pt,draw=mygray!50},
                    width=0.3\textwidth,
                ]
                \addplot[mark=square, color=myblue] table [x=x, y=ts36] {gru-globe.dat};
                \addplot[style={solid}, mark=diamond, color=mygreen] table [x=x, y=ts24] {gru-globe.dat};
                \addplot[mark=star, color=myorange] table [x=x, y=ts12] {gru-globe.dat};
                \addplot[mark=triangle, color=mycyan] table [x=x, y=ts6] {gru-globe.dat};
    
                \end{axis}
            \end{tikzpicture}            
        }
        \subfloat[T-GCN (radius 5)\label{fig:global:timeseries:tgcn}]{
            \begin{tikzpicture}
                \begin{axis}[
                    xlabel={Prediction horizon},
                    legend style={at={(0.5,1.1)},anchor=south,legend columns=-1},
                    xtick={1,4,8,12,16,20,24},
                    ytick={0.4, 0.45,0.5, 0.55,0.6, 0.65,0.7,0.8,0.9,1.0},
                    xmin=0,
                    xmax=25,
                    ymin=0.65, 
                    ymax=0.7,
                    grid=both,
                    major grid style={line width=.1pt,draw=mygray!50},
                    minor tick num=1,
                    every major grid/.style={line width=.1pt,draw=mygray!50},
                    width=0.3\textwidth,
                ]
                \addplot[mark=square, color=myblue] table [x=x, y=ts36] {tgcn-local-radius5-globe.dat};
                \addplot[style={solid}, mark=diamond, color=mygreen] table [x=x, y=ts24] {tgcn-local-radius5-globe.dat};
                \addplot[mark=star, color=myorange] table [x=x, y=ts12] {tgcn-local-radius5-globe.dat};
                \addplot[mark=triangle, color=mycyan] table [x=x, y=ts6] {tgcn-local-radius5-globe.dat};            

                \legend{36, 24, 12, 6}
    
                \end{axis}
            \end{tikzpicture}
        }
        \subfloat[Conv-LSTM (radius 5)\label{fig:global:timeseries:conv-lstm}]{
            \begin{tikzpicture}
                \begin{axis}[
                    xlabel={Prediction horizon},
                    xtick={1,4,8,12,16,20,24},
                    ytick={0.4, 0.45,0.5, 0.55,0.6, 0.65, 0.69,0.7, 0.71, 0.72, 0.73, 0.74, 0.75,0.8,0.9,1.0},
                    xmin=0,
                    xmax=25,
                    ymin=0.71, 
                    ymax=0.75,
                    grid=both,
                    major grid style={line width=.1pt,draw=mygray!50},
                    minor tick num=1,
                    every major grid/.style={line width=.1pt,draw=mygray!50},
                    width=0.3\textwidth,
                ]
                \addplot[mark=square, color=myblue] table [x=x, y=ts36] {conv-lstm-local-radius5-globe.dat};
                \addplot[style={solid}, mark=diamond, color=mygreen] table [x=x, y=ts24] {conv-lstm-local-radius5-globe.dat};
                \addplot[mark=star, color=myorange] table [x=x, y=ts12] {conv-lstm-local-radius5-globe.dat};
                \addplot[mark=triangle, color=mycyan] table [x=x, y=ts6] {conv-lstm-local-radius5-globe.dat};
    
                \end{axis}
            \end{tikzpicture}
        }
        \caption{Model performance at a global scale with different time-series length
            $ts \in \{1, 3, 6, 12, 24, 36\}$ periods (8-days).}\label{fig:global:timeseries}
    \end{figure*}

    In this section, we study the effect of training global models with longer or shorter
    input time series, with or without spatially accumulated information. Figure~\ref{fig:global:timeseries} shows the performance (AUPRC) of the models when 
    trained with four different time-series length $ts \in \{6, 12, 24, 32\}$. For each such
    length and each forecasting window size $tw \in \{1, 2, 4, 8, 12, 16, 20, 24\}$ we
    trained a separate model. 

    Figure~\ref{fig:global:timeseries:gru} contains the results for the GRU models.  
    The first important observation is that the length of the time series is important in order 
    to increase the predictive capabilities of the model. Models which work on input data 
    ranging up to 36 periods (8-day) in the past have better and more stable predictive capabilities and 
    depend less on the sub-seasonality of the input data. This sub-seasonality effect can be observed by the fact 
    that with length 6 time series, the models with target horizons 20 and 24 periods
    behave slightly better than the models with horizons 12 and 16 periods. This effect is 
    not present when the time-series length is 36 periods. In this case, as expected better 
    performance is achieved when the target horizon is a single 8-day period.    

    Adding a spatial component to the model while retaining the long-range temporal dependencies 
    seems to improve the prediction instability. 
    Figure~\ref{fig:global:timeseries:tgcn} show results using the temporal GCN. 
    The graph is constructed 
    using a radius $r=5$ meaning that the spatial grid-graph has $11 \times 11$ vertices. Each vertex 
    is connected to its $9$ closest neighbors including itself. We also experimented with other values 
    of $r$ but models with $r=5$ showed slightly better results. 
    The new models which capture both spatial and temporal aspects improve both the stability of the results, 
    especially for smaller timeseries lengths, but also achieve higher AUPRC. The best GRU model ranged
    from $\approx 0.61$ (target period 24) up to $\approx 0.64$ (target period 1) while the TGCN ranged
    from $\approx 0.66$ (target period 24) up to $\approx 0.69$ (target period 1). What is also interesting 
    is that the spatial dimension seems to compensate, up to a point, the losses of a smaller timeseries length. 
    When the timeseries length is 12 months, the models up to a target horizon of $8$ 8-day periods behave similarly to
    the models with larger timeseries length, i.e. 24 and 36.    

    Figure~\ref{fig:global:timeseries:conv-lstm} contains different Conv-LSTM models. The effect here 
    is stronger. Models with timeseries length 12 (or even 6) achieve identical performance as models
    trained with larger timeseries.

\subsection{Radius ablation}

The radius parameter $r$ affects the receptive field of both the Conv-LSTM and T-GCN models. 
    In both cases the models receive data in a spatial grid of size $(2r+1) \times (2r+1)$. 
    When $r=0$, the situation deteriorates to the case of the GRU model which has access to data
    from a single location.
    For larger $r>0$ the model also receives data from nearby 
    locations. Table~\ref{tbl:radius:ablation} contains the performance of separately trained 
    Conv-LSTM models for different values $r>0$ and different prediction horizons. 
    The results demonstrate that the performance gain from a larger radius, quickly saturates.
    Recall that for $r=5$ the spatial grid has size $11 \times 11$, which is already quite large 
    in order to predict only the central location.

\begin{table}[!t]
        \centering
        \begin{tabular}{|c|c|c|c|c|c|c|c|}
        \hline \multicolumn{8}{|c|}{ Conv-LSTM (AUPRC) } \\
        \hline \multirow{2}{*}{ \textbf{target} } & \multicolumn{7}{c|}{ $\pmb{r}$ (radius) } \\
        \cline{2-8} 
        & \textbf{1} & \textbf{2} & \textbf{3} & \textbf{4} & \textbf{5} & \textbf{6} & \textbf{7} \\
        \hline 1  &   0.726 &  0.736  &  0.741 &  0.742 &  0.745 &  0.744 &  0.743\\
        \hline 2  &   0.720 &  0.730  &  0.736 &  0.739 &  0.735 &  0.739 &  0.739\\
        \hline 4  &   0.714 &  0.720  &  0.725 &  0.728 &  0.732 &  0.734 &  0.735\\
        \hline 8  &   0.708 &  0.715  &  0.722 &  0.724 &  0.728 &  0.730 &  0.731\\
        \hline 12 &   0.703 &  0.716  &  0.721 &  0.724 &  0.722 &  0.725 &  0.729\\
        \hline 16 &   0.702 &  0.715  &  0.723 &  0.726 &  0.726 &  0.726 &  0.726\\
        \hline 20 &   0.707 &  0.716  &  0.726 &  0.725 &  0.724 &  0.729 &  0.727\\
        \hline 24 &   0.710 &  0.718  &  0.725 &  0.727 &  0.724 &  0.725 &  0.729\\        
        \hline
        \end{tabular}
        \caption{Performance of different Conv-LSTM models for different values of the 
        radius parameter $r$. Timeseries length is always $36$. 
        }
        \label{tbl:radius:ablation}
    \end{table}

    \subsection{Discussion \& Conclusions}

Our findings underscore the importance of considering longer time-series data for improving predictive capabilities. 
    We observed that models trained with longer time series, encompassing up to 36 periods (equivalent to an 8-day window), demonstrated superior predictive performance
    and stability compared to shorter time-series lengths.
    Particularly, models with longer time-series lengths exhibited reduced dependency on sub-seasonal variations in input data, leading to more robust predictions across
    varying forecasting horizons. Additionally, the incorporation of a spatial component, further enhanced prediction stability and achieved higher AUPRC, indicating the efficacy of integrating spatial information to capture complex spatiotemporal dynamics inherent in fire prediction tasks.

    Furthermore, our study sheds light on the limitations and potential avenues for future research in enhancing the predictive capabilities of deep learning models for fire 
    prediction. Despite the improvements observed with longer time-series data and spatially-aware models, we identified a plateau when trying to predict over long-range horizons. 
    When the prediction horizon is larger than 12 8-days periods, the model seems to fallback to the average mean cycle performance.
    This might indicate that the local receptive field of is not enough and we should add long-range teleconnections and additional global information 
    in order to go further in time. 
    In contrast to ConvLSTM, GNNs exhibit notable flexibility in connecting remote nodes, and taking into consideration global information. This ability provides a more
    comprehensive understanding of intricate patterns and dependencies, making GNNs potentially particularly relevant in scenarios where global context plays a pivotal role.

    \section*{Acknowledgment}
    This work has received funding from the SeasFire project, which deals with 
    "Earth System Deep Learning for Seasonal Fire Forecasting" and is funded by the
    European Space Agency (ESA) in the context of ESA Future EO-1 Science for Society Call.

    \bibliographystyle{plain}
    \bibliography{bibliography}

\end{document}